# Artificial Dendritic Cells: Multi-faceted Perspectives

**Julie Greensmith, Uwe Aickelin**

University of Nottingham, School of Computer Science, Nottingham, NG8 1BB, UK,
{jqg,uxa}@cs.nott.ac.uk

**Abstract:** Dendritic cells are the crime scene investigators of the human immune system. Their function is to correlate potentially anomalous invading entities with observed damage to the body. The detection of such invaders by dendritic cells results in the activation of the adaptive immune system, eventually leading to the removal of the invader from the host body. This mechanism has provided inspiration for the development of a novel bio-inspired algorithm, the Dendritic Cell Algorithm. This algorithm processes information at multiple levels of resolution, resulting in the creation of information granules of variable structure. In this chapter we examine the multi-faceted nature of immunology and how research in this field has shaped the function of the resulting Dendritic Cell Algorithm. A brief overview of the algorithm is given in combination with the details of the processes used for its development. The chapter is concluded with a discussion of the parallels between our understanding of the human immune system and how such knowledge influences the design of artificial immune systems.

## 1. Introduction

The human immune system (HIS) is a decentralised, robust and error tolerant system which consists of a plethora of interacting cells. This system provides protection from invading entities such as bacteria and regulates numerous bodily functions. Immunology, the study of the human immune system, encompasses multiple levels of abstraction. For the past 100 years immunology has been a reductionist science, concentrating on the precise mechanisms involved in the relationship between immune-related molecules and cells. More recently [Cohen07] immunologists are examining such components from a systemic perspective. The exact purpose of the HIS still remains elusive, though current thinking within immunology is that it provides a combination of *protection* and *regulation*. Protection involves the rapid detection of invading microorganisms termed *pathogens*, their subsequent removal from the body and the process of repair following pathogenic infection. Regulation via the immune systems involves the maintenance of a constant internal environment, namely the homeostasis of bodily processes. This includes temperature, acidity levels, growth of blood vessels, regulation of inflammatory processes and tolerance to self-cells.

As computer scientists, our interest in the immune system is as a protection system as natural parallels can be drawn between natural pathogens and threats to computer systems, such as internet based 'viruses' and 'worms'[Forrest94]. To use the immune system as inspiration for computer algorithms, the construction of immune-inspired models is performed at numerous levels of abstraction, including molecular signaling networks and models of cell. These concepts are translated into an algorithm or system through processes of abstraction and modeling.

The creation of artificial immune systems (AISs) involves the translation of basic immunological models into feasible algorithms. This requires careful modeling of immune inspired features. To achieve this successfully, it is recommended that the desired immune components are modelled at various levels of abstraction then transformed into an algorithm using a similar multi-scale ethos.The choice of functions abstracted from the natural system is heavily influenced by the methods used in experimental immunology as this limits our understanding of the immune system. Three different levels of abstraction are commonly used including the molecular level, cellular level and systemic level, with the majority of research focusing on the molecular level. Such trends within immunology influence the manner by which AISs are created with most using models of molecular interactions in terms of binding between molecules [deCastro02].

The Dendritic Cell Algorithm (DCA) is an example of an immune inspired algorithm developed using a multi-scale approach. This algorithm is based on an abstract model of *dendritic cells* (DCs). The DCA is abstracted and implemented through a process of examining and modeling various aspects of DC function, from the molecular networks present within the cell to the behaviour exhibited by a population of cells as a whole. Within the DCA information is granulated at different layers, achieved through multi-scale processing. This differs from the standard view of granular computation [Bargiela03] as such information granules do not have an explicit fuzzy component or membership function. However, their processing is performed in a similar multi-level manner and across multiple time scales forming a diverse set of information granules. Input data is in the form of two different input streams, which are combined and correlated across variable time windows. In addition such AIS algorithms are inherently human-centric in their development. They are based on a foundation of how the immune system is perceived through immunological experimentation. This ultimately forms the abstract biological model underpinning the function of immune inspired computational systems.

In this chapter we use the parallels drawn between natural DCs and the artificial population of DCs used in the DCA to illustrate the principles behind multi-scale algorithm development. The aim of this chapter is to show how such abstraction can be achieved and to stress the importance of understanding a system from multiple perspectives to produce systems that encompass several layers of information granularity. In Section two an overview of the relevant immunology is presented and the development process of multi-signal models is outlined. In addition we present an overview of the AIS algorithms developed for computer security and optimisation and draw parallels with human-centric developments in immunology. Section three introduces the DCA as a multi-resolution algorithm. Section four provides a brief description of an implemented DCA highlighting signal and antigen processing as granular computation, and Section five continues the discussion of the DCA in the context of human centric development. Finally conclusions are drawn regarding the relationships between immunology, AIS and the lessons learned from the developmental process used to create the DCA.

## *2. Background*

### *2.1 Human Immune System*

The human immune system (HIS) is vast, containing in excess of 10 million cells. There is no archetypal "immune cell" akin to neurones in the central nervous system. Instead the HIS is an abstract concept, a name imposed by immunologists for a collection of cells whose function is within the remit of protection and regulation. The HIS is classically subdivided into two distinct branches: the *innate* and the *adaptive* systems. The innate system is evolutionarily the oldest immune component and its role to provide a rapid response on detection of specified molecules within the body [Murphy08].

Innate cells include macrophages, natural killer cells and dendritic cells, which perform initial pathogen detection by instructing the immune system of damage and clear the surrounding tissue of any debris. Over the evolution of the species, the immune system has acquired the knowledge of which molecules indicate the presence of pathogens. Immune cells are equipped with receptors (surface bound proteins) armed to detect such molecules. These receptors are present in great number on the cells of the innate immune system. The repertoire of pathogenic recognition receptors (termed *pattern recognition receptors)* is fixed once the genome of an individual is encoded. This implies that the innate immune system cannot adapt to novel threats over the lifetime of the individual - an important task given the fact that pathogens are constantly evolving.

To keep pace within this biological arms race, the HIS also contains a population of cells that are able to dynamically restructure their receptors to adapt to new threats. Such cells of the adaptive immune system, namely B-cells and T-cells, have the ability upon cloning to reorganise the molecules of their pathogen detectors (termed *variable region receptors*) to attempt to adapt to new threats. It is the combination of the rapid response of the innate immune system, coupled with the dynamic modifications of the adaptive immune system that provides sufficient protection to ensure the survival of our species.

The current thinking of immunologists heavily influences the manner by which we construct AISs. The inspiration used as the basis of such algorithms is derived not from the immune system itself, but from human abstractions of how we believe the immune system to function. Therefore here we introduce the basic trends in immunology over the past 100 years and comment on how various human-centric streams of research in immunology has influenced the field of AIS.

In 1891, Paul Ehrlich and his colleagues [Silverstein05] postulated that the human defense mechanism against pathogens revolved around the generation of immunity through the production of antibodies. He showed that these generated antibodies are specific to the pathogen or toxin being targeted. From his perspective a paradox existed termed *horror autotoxicus* - the immune system has to ensure that invaders are controlled and deleted before an infection spreads without responding to or damaging its own cells. Following Ehrlich's work, the *clonal selection principle* was developed where the immune system is postulated to have the ability to respond to proteins - termed antigen - which do not belong to 'self' and to target antigens belonging to 'nonself i.e. pathogenic proteins. This formed a major constituent of a theory known as central tolerance and is shown in Figure 1 as the "one-signal model".

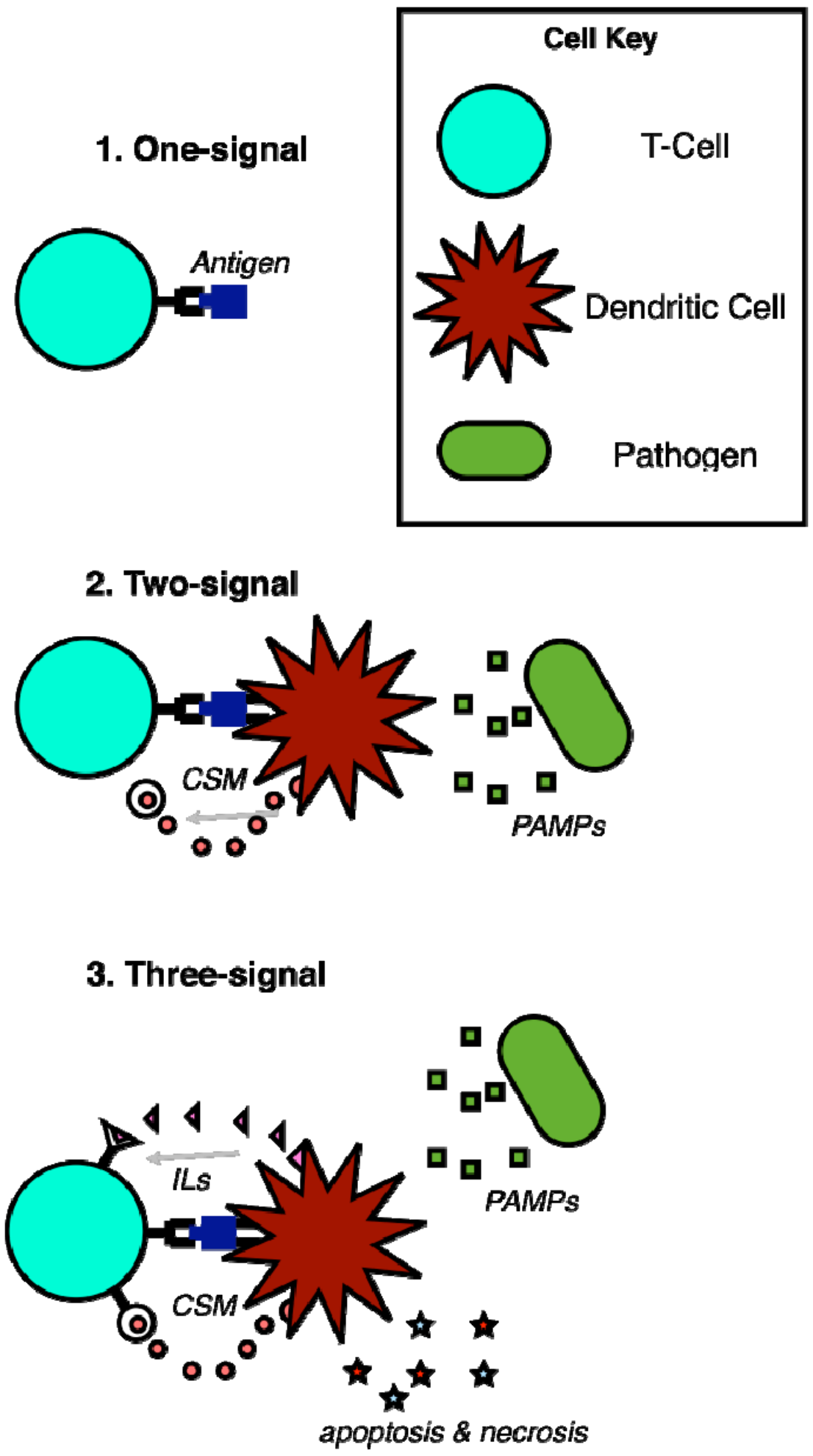

**Figure 1** Abstract representation of the three multiple signal models developed in immunology, including the original one signal model, the costimulation driven two signal model where the involvement of pathogens was not understood until after the model was explored, and finally the three signal model which also includes danger signals.

As the 20th century progressed, T-cells were characterised in addition to the antibody producing B-cells. In 1959, Lederberg proposed the principle of negative selection. He established the link between foetal development and the generation of tolerance to self-antigen. It is shown that in infancy, newly created T-cells are presented a sample of self-antigen. T-cells are deleted if they displaying a receptor protein which matches self-antigen with a sufficiently high binding affinity. This results in a population of T-cells acutely tuned to respond to non-self entities.

However, this response to non-self is not always an observable fact and numerous noteworthy exceptions have been discovered [Murphy07]. Four main problems have arisen questioning the credibility of central tolerance and 'self-nonself' as the central dogma of immunology.

- Vaccinations and immunisations require adjuvants (bacterial detritus) despite the vaccination containing non-self particles;
- What the body classes as self changes over time for example in pregnancy;
- Our guts are host to colonies of bacteria which serve a symbiotic function forming the gut flora, without which we are prone to severe intestinal infections and inflammation;
- The immune system can behave inappropriately and attacks its host in the form of autoimmune diseases such as multiple sclerosis, rheumatoid arthritis and inflammatory bowel disorders, in addition to the generation of allergy to benign particles such as pollen.

The first major modification to the classical one-signal model is the addition of a secondary pathway for the activation of adaptive immune cells. This is termed *costimulation* and has been shown as a requirement for the full activation of T-cells, forming the two-signal model shown in Figure 1. Even if an antigen and T-cell bind sufficiently well, a costimulation signal is required in order for the activation of the T-cell to effector function. In order to bind to antigen, a T-cell must be 'presented' the antigen by a cell of the innate immune system, known as an *antigen presenting cell (APC)* such as DCs.

It is thought that for a T-cell to become activated it must be first presented its antigen by an APC in conjunction with molecules termed co-stimulatory molecules (CSM). Initially it was undetermined as to what causes APCs to express such molecules. Janeway [Janeway89] postulated that APCs produced CSMs in response to the detection of bacterial sugars, known as PAMPs - *pathogen associated molecular patterns.* These molecules are exclusively produced by pathogens as the name suggests and hence act as a signature of bacterial presence in the body. This is a 'two-signal model' (Figure 1) as the T-cell is given two signals, CSM and antigen.

This theory explains the need to add bacterial detritus to immunisations, and also the lack of response to changing self-proteins, as they do not have PAMPS. However, this theory alone cannot explain the lack of response by the immune system to the 'friendly' bacteria in the gut or the phenomenon of auto-immunity to which no pathogens are present.

One of the most recent models is the "danger theory" which incorporates a third signal. Matzinger [Matzinger94] proposed that in addition to the requirement for antigen and CSMs, T-cells also require a particular type of *interleukin,* a messenger molecule, from the APC to promote full T-cell activation. The danger theory postulates that this particular interleukin is produced by the APC in response to exposure to tissue damage. This ‘three-signal model’ is shown in Figure 1.

It is thought that the presence of something like a bacterial colony in an otherwise healthy piece of tissue will cause the tissue cells to die unexpectedly. As a result of such cell death, termed *necrosis*, the inner constituents of the tissue cells are subject to a rather chaotic degradation process. DCs in particular are shown to increase production of the relevant interleukin, IL-12, upon receipt of such indicators of cell damage.

Conversely, cells can die as part of a normal regulatory process, termed *apoptosis.* DCs exposed to the signals of apoptosis themselves produce a different kind of signal, termed IL-10. Instead of activating the T-cell, production IL-10 by DCs causes T-cell deactivation. Through DCs producing varying amounts of IL-12 versus IL-10, the T-cell is given final confirmation whether to respond to the presented antigen or to become tolerant to its presence.

Research continues in immunology to find further plausible mechanisms of immune activation. Recently, a new type of T-cell, a Th17 cell has come to the fore using a fourth signal expressed by DCs. The mechanism of action still remains unclear, though it appears that this cell is stimulated without the third interleukin signal and in the presence of a fourth signal known as TGF-β. These discoveries show that no matter what the current state of the art, such models will be continually updated and improved as we develop increasingly sophisticated techniques for the study of the function of the HIS, leading in perpetual development of AIS based on these new discoveries.

To summarise, multiple-signal models of T-cell activation have dominated much of immunology for the past century. This basic model has been subject to much debate and numerous additions incorporating different molecular activating and suppressing signals in addition to the binding of T-cell to antigen. Understanding the basics of immunology is the initial step in creating AIS algorithms. In the next section we discuss how AISs have developed in a similar manner to the multiple signal models presented in this section.

## *2.2 Artificial Immune Systems (AISs)*

AISs are computer systems and algorithms inspired by the function of the HIS. There are numerous parallels in the pathway of development of AISs. As with immunology, AIS also began by using the self-nonself principles of negative and clonal selection to create the Negative Selection Algorithm, which was used primarily for applications within computer security. Clonal selection is used in a variety of immune algorithms including AIRS, which has proven to be a competitive multi-class classification system.

In comparison with other bio-inspired computing paradigms, AISs are relatively young. Forrest *et al.* first implemented negative selection in 1994 [Forrest94], based on the T-cell centric one-signal model. Exposition and exploration of this algorithm dominated the field of AIS for the following decade. The idea behind this principle is appealing to computer security - the notion of creating a computer immune system to detect computer viruses is naturally appealing as a metaphor. This algorithm uses self-nonself principles, creating a set of randomly generated 'detectors' tuned via a mechanism of profiling of normal behaviour, selection of detectors which deviate from normal. This results in a detector set tuned to only respond to 'non-self' or anomalous strings.

While negative selection generated much interest in AIS, the algorithm itself has been shown to have a number of shortcomings. The nature by which the detectors are generated relies on the initial creation of a sufficient amount of detectors to cover the potential self-nonself feature space. Obviously, as the dimensionality or size of this feature space increases, the number of detectors required to fully cover such space increases exponentially. This has been proven both experimentally [Kim01] and theoretically [Stibor06]. In addition to such scaling problems, the algorithm also is prone to the generation of false alarms or false positives. These type1 misclassification errors are thought to arise partially due to the 'one-shot' style of learning, and the fact that it is difficult to accurately represent what is 'normal' within a single bit-string [Stibor05]. Despite numerous attempts to remedy this challenge with thorough investigations of the representation [Zhou06], this algorithm does not produce results similar in calibre to that observed by the HIS.

Consequently AIS researchers have incorporated an ever-increasing amount of underlying immune-inspiration in an attempt to improve such algorithms. For example, the incorporation of a second signal was first proposed by Hofmeyr [Hofmeyr99] and implemented by Balthrop [Balthrop05], where it was shown to reduce the rates of false positives in numerous cases. As with immunology, AIS has continued to add signals to its underlying models in much the same manner as immunologists have over the past century.

Aickelin *et al.* proposed a novel approach to the development of AIS [Aickelin03] centered in the incorporation of the danger theory to AIS. Two streams of research resulted from this proposition, one including Janeway's infectious nonself model and the other resulting in the creation of the Dendritic Cell Algorithm. Both algorithms are applied with success to the detection of network intruders, encompassing a variety of problems within such fields [Greensmith06, Twycross06].

The augmented two-cell model was implemented by Twycross and Aickelin [Twycross06] and while it was never explained as incorporating a PAMP signal (it is expressed as a 'danger signal' in their literature) it is indeed incorporating a secondary signal to a process which also requires the selection of a T-cell along with the use of an APC to provide the second signal. The second signal was derived from data out of range of characterised 'normal' data.

The developments of AIS outlined above do not focus on the development of clonal selection and idiotypic network based systems, as they do not have sufficient relevance to the development of the DCA. However, the interested reader is advised to refer to Timmis and DeCastro [Timmis02] for further details.

The development of AIS for uses within computer security in particular have inherent parallels with dogma in immunology as summarised in Figure 2. This can be attributed to the fact that AIS researchers are improving their relationships with practical immunologists as interdisciplinary collaborations become increasingly prevalent within computer science and the life sciences. This was indeed the case for the 'Danger Project' resulting in the development of the DCA. This is corollary to the fact that techniques in immunology have developed to such a level where quite detailed models can be constructed as the knowledge base expands regarding the actual function of the HIS.

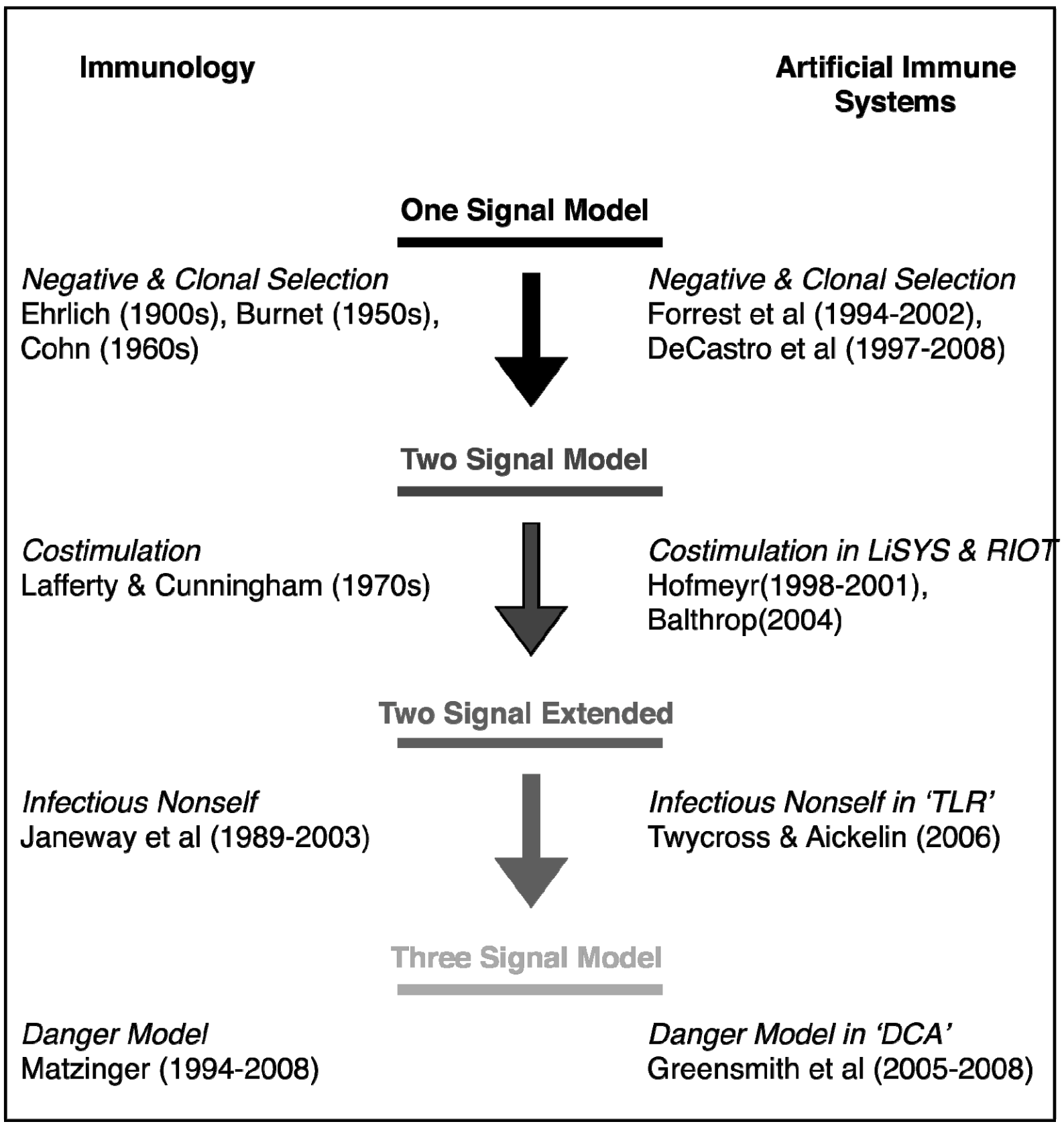

**Figure 2** The parallel development of immunology and subsequently, artificial immune systems. Given the trend in artificial immune systems to work increasingly closely with immunologists, we expect that this trend will continue for the forseeable future within this field.

## 3. Overview of the Dendritic Cell Algorithm (DCA)

In this section we give an overview of the DCA and its underlying immune inspiration. Metaphorically, DCs are the crime-scene investigators of the HIS, traversing the tissue for evidence of damage - namely the signals of apoptosis and necrosis, and for potential culprits responsible for the damage, namely antigen. More information regarding the function of natural DCs can be found in an excellent review by of the field by Lutz and Schuler [Lutz02] with a distilled version for computer scientists presented in [Greensmith07].

The DCA is derived from an abstract model of DC biology resulting in an anomaly detection algorithm that provides robust detection and correlation. Different cells process input data mapped as 'signals' acquired over different time periods. This generates individual 'snapshots' of input information that are subsequently correlated with antigens. The DCA is described in greater technical detail in numerous sources including Greensmith *et al.* [Greensmith06, Greensmith08a] and in the corresponding PhD thesis [Greensmith07].

The process of creating an algorithm such as the DCA is nontrivial, involving multiple stages of development and requires the performance of cross-disciplinary research in conjunction with immunologists. Within the framework of the Danger Project [Aickelin03], practical immunologists conducted parallel research which filled gaps in knowledge to assist in the creation of the most accurate models possible. In this section a high level description of the algorithm is provided for illustrative purposes.

The DCA is a population based algorithm, with each artificial cell acting as an agent within the system. To achieve the incorporation of our abstract model of DC function two levels of abstraction are used, namely the internal mechanisms of the cell and the overall behaviour of the cell throughout its lifetime. As an algorithm it performs filtering of input signals, correlation between signals and antigen, and classification of antigen types as normal or anomalous. Two levels are explicitly modelled, namely the internal cell procedures and the behavioural state changes.

The internal cell procedures form the lowest level of abstraction used to dictate the behaviour of the artificial DCs. This comprises the collection of antigen data and the cumulative processing of the cells input signals. Input signals are transformed into cumulative output signals acquired over time. Signal data enters the system and is stored in an array. The cell uses these signal values each time the cell update function is called. Upon acquisition of the signal values each cell performs a weighted sum equation to combine the inputs three times to produce interim output values. These interim values are added to a final output value resulting in each cell producing three 'running total' output signals. Each input signal has a weight associated to transform the input values into the three interim values. The model of this process is represented in Figure 3.

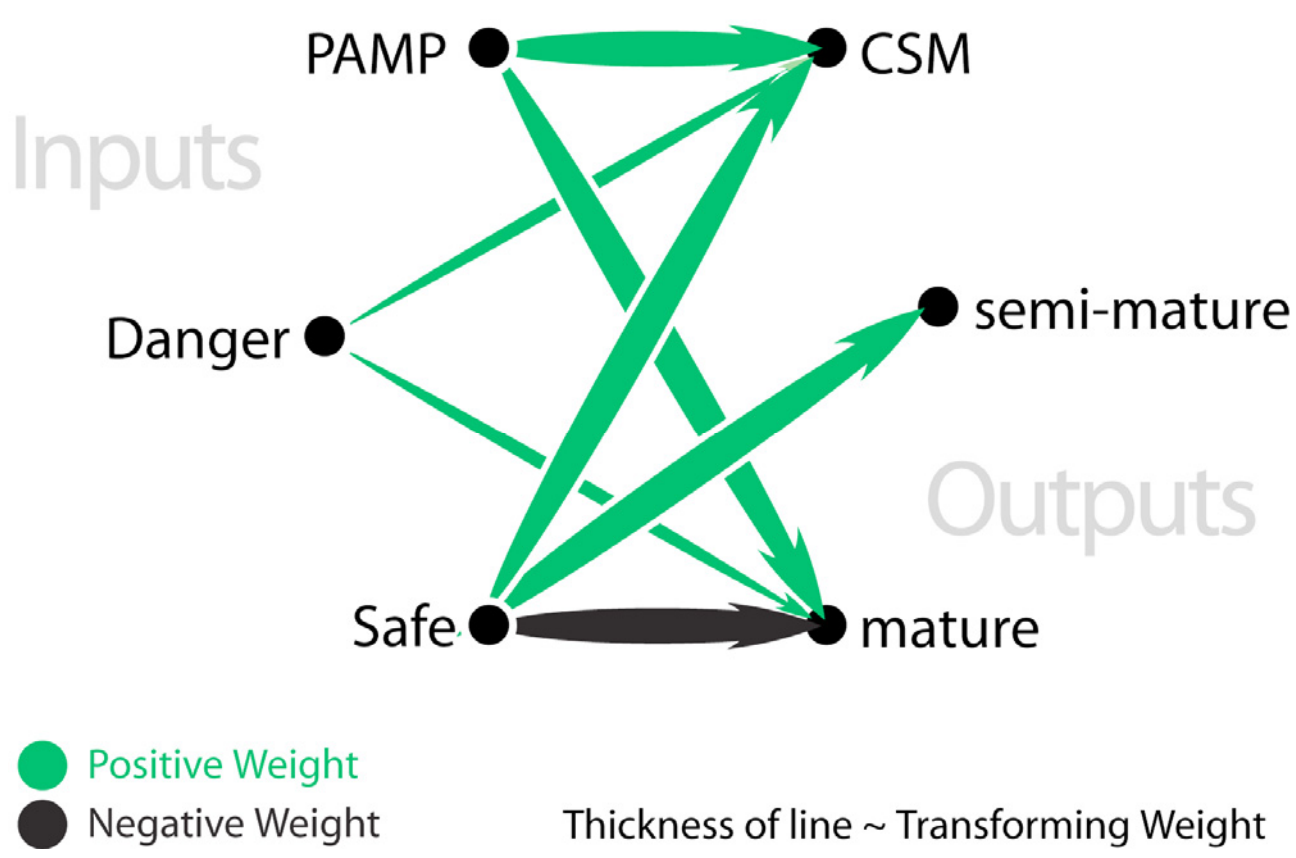

**Figure 3** Graphical representation of the signal processing used within each cell of the DCA. Each input signal per category is transformed to one of the three outputs. The weights used in this calculation are derived from a ratio discovered by our associate immunologists.

Each output signal is assessed each iteration. Three output signals are generated termed the costimulation value; the semi-mature output; and the mature output and their respective functions described in Table 1. The costimulation value is used to limit the lifespan of each individual cell within the DC population. Upon initialisation, each cell is assigned a threshold value, representing the lifespan of the cell. The cumulative costimulation value is assessed against the lifespan threshold each iteration. Once this threshold is exceeded, the cell is removed from the population, analysed and eliminated. Upon analysis, the remaining two values are assessed.

| Output signal | Function |
|---|---|
| Costimulatory signal | Assessed against a threshold to limit the duration of DC signal and antigen sampling, based on a migration threshold |
| Semi-mature signal | Terminal state to semi-mature if greater than resultant mature signal value |
| Mature signal | Terminal state to mature if greater than resultant semimature signal value |

**Table 1** Cumulative output signal functions for the three output signals of an artificial DC.

The behavioural component is summarised in Figure 4. This level of abstraction is used to define the state changes that appear to be so pivotal to the role of the DC in the HIS. In nature DCs change state to either mature or semi-mature at a certain point. In our abstract model the DCs have perceived sufficient information when they produce a particular receptor attracting the cell to the lymph node compartment.

The costimulatory value controls the initial state change from immature to either the semi-mature or mature state. The final state is determined by the greater of the two remaining values. If the value of the semi-mature output is greater then the cell is deemed semi mature, and the same process applies should the mature signal be greater.

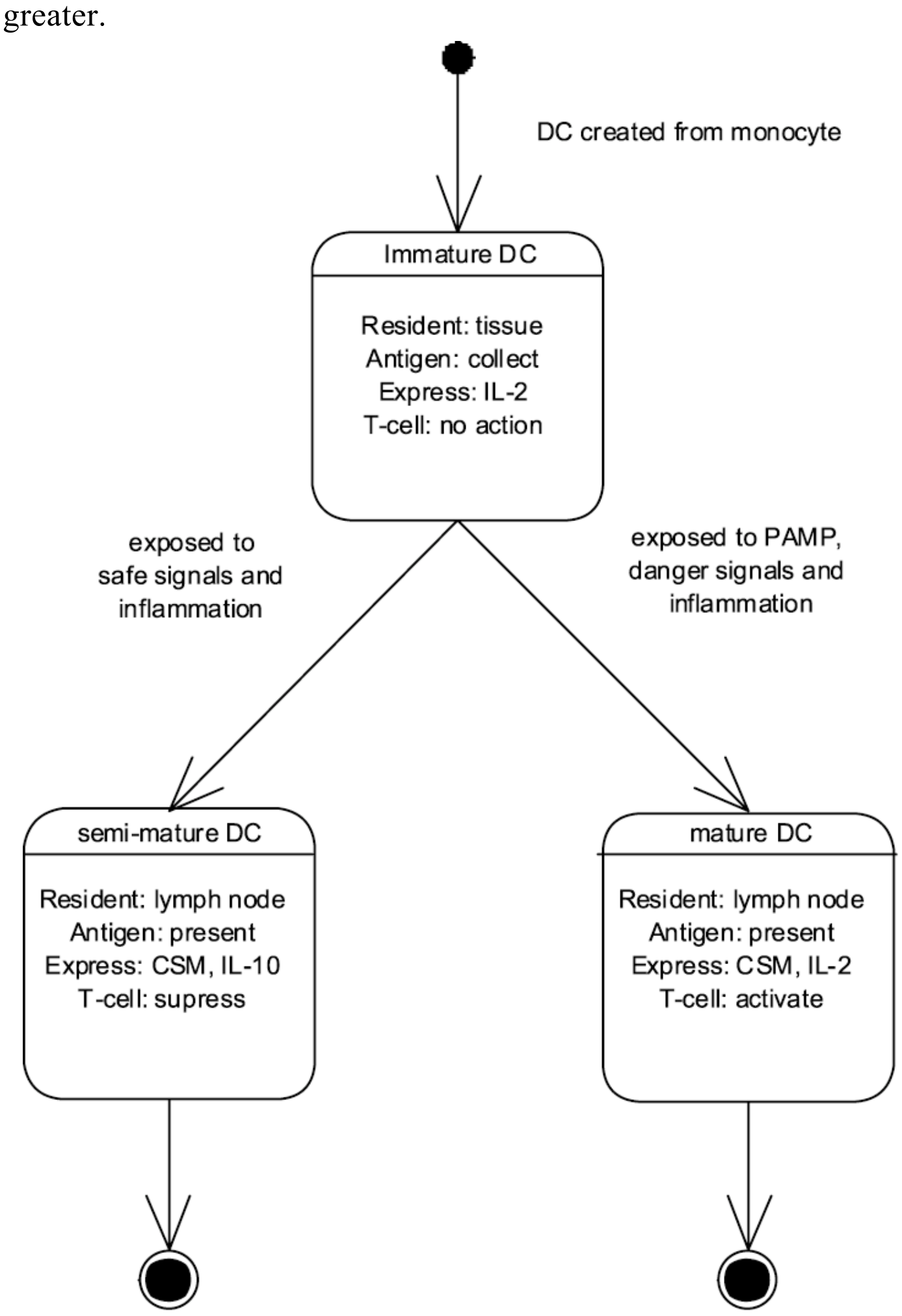

**Figure 4** UML state chart representing the abstract model of an individual DC.

Each time input signals are received an antigen may also be collected (unless no antigen are generated at that timepoint). All antigen collected by the cell over its lifetime are 'presented' in conjunction with this context value at this analysis stage. The antigen-plus-context is used to assess the anomalous nature of such antigen. Antigen are not modified in any way by the DC, but are collected from the antigen vector and stored until presentation. The manner by which the antigen are stored has varied between the various versions of the algorithm, though this is not thought to affect the resulting performance of the algorithm.

A minimum number of ten cells are required to perform processing [Greensmith08]. The multiplicity of cells means that the algorithm uses a consensus decision generated across the population to make decisions. The output of the algorithm is an anomaly score for each antigen type, to which a threshold can apply to give a definite class prediction. Due to the time-sensitive nature of the algorithm, it is not particularly suited to randomly ordered data but is shown to have useful and robust properties when applied to challenging real-time applications [Greensmith07]. The abstract principles outlined in this section are further elaborated upon in Section 4 to demonstrate how this algorithm works in practice.

## 4. Implementing the DCA

In Section 3 a high level overview of the DCA is given. In this section a more detailed algorithmic description is given. The purpose of a DC algorithm is to correlate disparate data-streams in the form of antigen and signals and to label groups of identical antigen as 'normal' or 'anomalous'. The DCA is not a classification algorithm, but shares properties with certain filtering and sorting techniques. This is achieved through the generation of an anomaly coefficient value, termed the MCAV. The labeling of antigen data with a MCAV coefficient is performed through correlating a time-series of input signals with a group of antigen. The signals are pre-normalised and pre-categorised data sources based on snapshots of preliminary experimental data, which reflect the behaviour of the system being monitored. Categorisation of the signals is based on the four signal model based on PAMP, danger, safe and inflammation signals. The co-occurrence of antigen and high/low signal values forms the basis of categorisation for the antigen data. The primary components of a DC based algorithm are as follows:

1) Individual DCs with the capability to perform multi-signal processing
2) Antigen collection and presentation
3) Sampling behaviour and DC maturation state changes
4) A population of DCs and their interactions with signals and antigen
5) Incoming signals and antigen, with signals pre-categorised as PAMP,

danger, safe or inflammation

6) Multiple antigen presentation and analysis using 'types' of antigen
7) Generation of anomaly coefficient for various different types of antigen

Each cell in the population acts as an agent and has a set of instructions, which are performed every *cell update* iteration. Control of the frequency and nature of cell updates is specific to the instance of the algorithm's implementation as is the rate of signal sampling and the number of antigen collected per iteration. Diversity is generated within the DC population by initiation of migration of the DCs at different time points i.e. the cessation of data sampling. This creates a variable time window effect throughout the DC population, which adds robustness to the system, and segments signals and antigen into variable information granules which underpin the functioning of the algorithm.

Each time a cell is updated the input signals are processed and added to the cell's internal values to form a set of cumulatively updated output signals in addition to the collection of antigen data items. The DCs are assigned one of three states at any point in time, namely immature, semi-mature or mature. Initially the cells are all assigned the 'immature' state label. Upon the receipt of sufficient signal values to initiate a process termed maturation, the cell can transform to either the semi-mature or the mature state. The differences in the semi-mature and mature state are controlled by a single variable, determined by the relative differences between two output signals produced by the DCs. If, over its lifespan, the cell accumulates predominantly safe signals, the cell is classed as semi-mature, otherwise it is assigned the mature status. Whilst in the immature state, the DC has three functions, which are performed each time a single DC is updated:

1) *Sample antigen*: the DC collects antigen from an external source and places the antigen in its own antigen storage data structure.
2) *Update input signals*: the DC collects values of all input signals present in the signal storage area
3) *Calculate interim output signals*: at each iteration each DC calculates three temporary output signal values from the received input signals, with the output values then added to form the cell's cumulative output signals.

Signal processing is performed via a weighted sum equation, bypassing the modelling of any biologically realistic gene regulatory network. A simple weighted sum equation is used in order to reduce any additional computational overheads, as the primary purpose of this algorithm is to perform anomaly detection in near to real-time. The crucial component of this procedure is the ability of the user to map normalised input data to one of the four categories of input signal (PAMP, danger, safe and inflammation). The general form of the signal processing equation is:

$$O_p = (P_w \Sigma_i P_i + D_w \Sigma_i D_i + S_w \Sigma_i S_i)\ (1+I) \qquad \forall\ p\ ,$$

where $P_w$, $D_w$ and $S_w$ are assigned weights, $P_i$, $D_i$ and $S_i$ are the input signal values of category PAMP (P), danger (D) or safe (S) for all signals ($i$) of that category for all output signals $p$, assuming that there are multiple signal sources per category. In this equation, the term $I$ represents the inflammation signal. This sum is repeated three times, once per output signal, which are then added to the cumulative output signals. Suggested ratios for the weights are given in Table 2 where input signals are represented as $j$ per category and outputs as $p$ per value. Each weight can be derived from two weights directly assigned to the PAMP signals ($W1$ and $W2$). The actual values used for the weights can be user defined, though the relative values determined from biological experimentation are kept constant.

| | $i$ = 1, PAMP | $i$ = 2, Danger | $i$ = 3, Safe |
|---|---|---|---|
| $p$ = 1, costimulation | $W1$ | $W1$ / 2 | $W1$ * 1.5 |
| $p$ = 2, semi-mature | 0 | 0 | 1 |
| $p$ = 3, mature | $W2$ | $W2$ / 2 | $W2$ * -1.5 |

**Table 2** Derivation and interrelationship between weights in the signal processing equation, where the values of the PAMP weights are used to create the all other weights relative to the PAMP weight. $W1$ is the the weight to transform the PAMP signal to the CSM output signal and $W2$ is the weight to transform the PAMP signal to the mature output signal.

Each member of the DC population is assigned a context upon its state transition from immature to a matured state of either mature (context = 1) or semi-mature (context =0). Diversity and feedback in the DC population is maintained through the use of variable migration thresholds. The natural mechanism of DC migration is complex and not particularly well understood, involving the under and over production of numerous interacting molecules. Therefore we use a simple approximation of a thresholding mechanism using migration thresholds to assess if a DC has received sufficient information to present suitable context information along side the antigen collected during this sampling period.

Each DC in the population is assigned a “migration threshold value upon its creation. Following the update of the cumulative output signals, a DC compares the costimulatory signal value (CSM) with its assigned migration threshold value. If CSM exceeds the migration threshold, the cell ceases sampling input data and the resultant values and collected antigen are ‘presented’ for analysis. At this point the cell is reset (all internal values set to zero and antigen expunged) and returned to the sampling pool of cells.

The range of the migration thresholds assigned throughout the population is also a user-defined parameter. Previously random, Gaussian and uniform distributions have been used to provide the population with this diversity with respect to the range. We have used simple heuristics to define the limits of such ranges of threshold value, relating to the median values of the input signal data, and as a result are data-specific. The net result of this is that different members of the DC population 'experience' different sets of signals across a time window. If the input signals are kept constant, this implies that members of the population with low values of migration threshold present antigen more frequently, and therefore produce a tighter couple between current signals and current antigen. Conversely, DCs with a larger migration threshold may sample for a longer duration, producing relaxed coupling between potentially collected signal and context. This diversity ensures that the same information is processed in slightly different manners, resulting in noise tolerance to variation and conflict in the input data streams.

Once all data is processed or a specified number of antigen are presented (if the dataset is sufficiently large) the antigen and cell context values are collated to form the anomaly scores for each antigen type. The antigen data used with the DCA are an enumerated type variable, with multiple antigen of the same value forming a single antigen type. For example, a running process on a CPU has a process ID, and the antigen can be a representation of the process ID generated each time the process invokes a system call.

Antigens are collected by different DCs that have experienced different snapshots of signal data. Therefore to analyse an antigen type one must average the experience of the DC population for that particular type. The value we assign per antigen type is termed the *mature context antigen value* or *MCAV*. This is a real value between zero and one: the closer this value is to one the greater the probability that this antigen type is anomalous. The MCAV is the sum of the number of individual antigen presented in the mature context divided by the total number of antigen presented for a single antigen type. This forms an average context value for each antigen type calculated from information derived using the population dynamics of the algorithm. The creation of this value also adds robustness as it cancels out any errors made by individuals in the DC population. At the core of this algorithm is a combination of numerical signal data processed at the lowest level of granularity, correlated with enumerated type antigen at a higher level of abstraction, which when brought together results in a robust anomaly detection paradigm. A generic version of the DCA is shown in Figure 5.

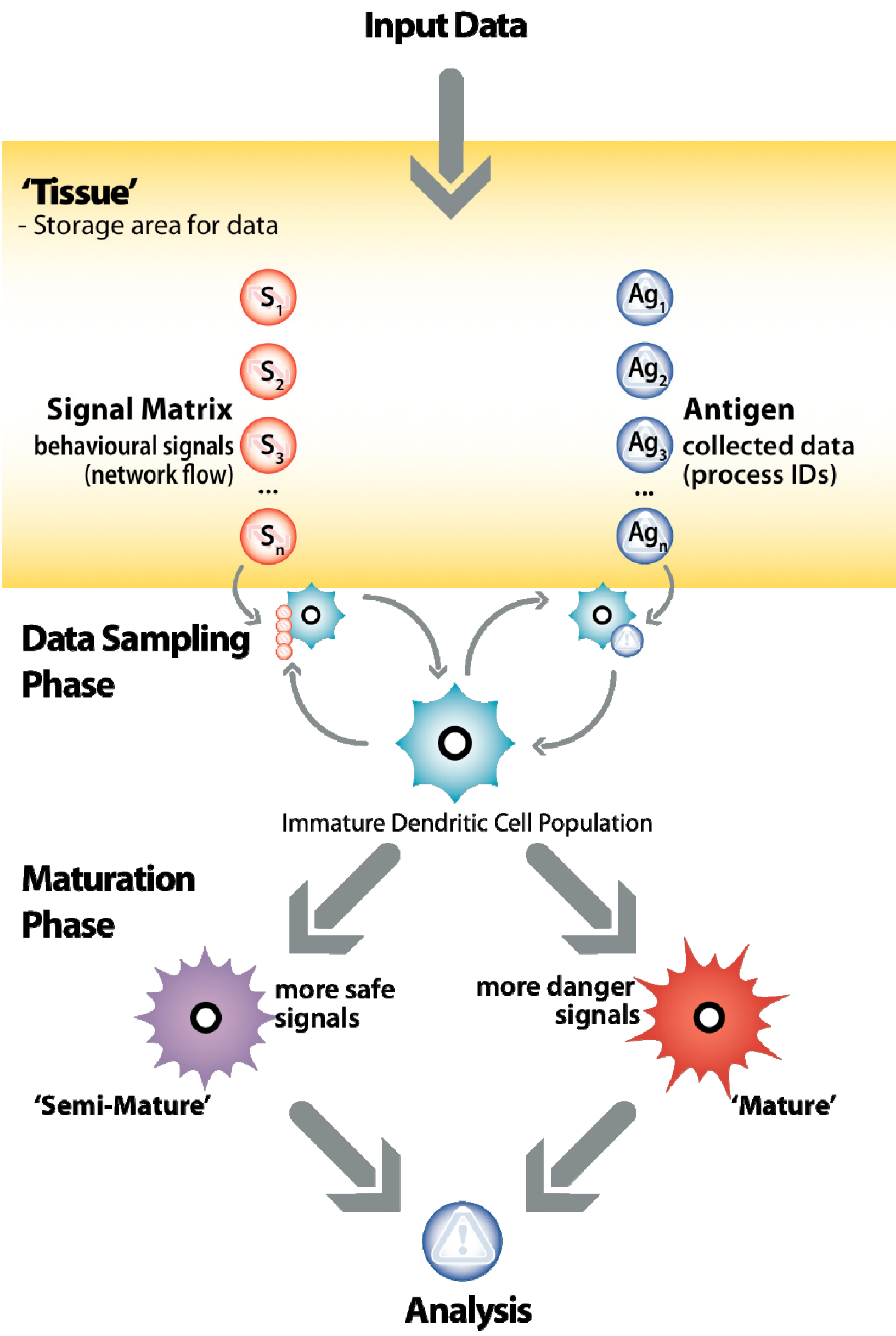

**Figure 5** A high level overview of the DCA as a system, with data flowing in to the signal matrix and antigen storage areas, and antigen types presented for analysis where the MCAV anomaly values are generated.

The DCA has been applied to numerous anomaly detection problems where signal and antigen mapping is possible. Such scenarios include the detection of port scans and internal intrusions. The detection of internal intrusions with the DCA formed a significant development for this algorithm, with extensive experimentation and analysis performed [Greensmith08a]. The dataset used was derived from monitoring a real host machine under a variety of experimental scenarios, such as emulating busy 'mid-morning' periods and performing scanning attacks under different network conditions. The objective of the insider experiments is to assess the DCA's performance when applied to the detection of slow and stealthy port scans. The antigen types are captured process IDs generated by the host machine each time a monitored process invokes a system call. The seven used signals are monitored from various system attributes of the monitored host:

1) PAMP1 : Number of ICMP destination unreachable errors received per second;
2) PAMP2 : Number of TCP Reset packets received per second;
3) Danger1 : Sending of network packets per second;
4) Danger2 : Ratio of TCP to all other packets per second;
5) Safe1 : Rate of change of sending network packets per second;
6) Safe2 : Average TCP packet size;
7) Inflammation: Presence of a remote root login.

In these experiments we show that the DCA has the ability to discriminate between the standard running processes on a monitored machine and an anomalous sustained port scan, performed by an emulated internal intruder. The results of this study also highlight a susceptibility of the algorithm to the 'bystander' effect, as a small number of false positives are generated to a normal process if it is equally as active as an anomalous process at exactly the same time. For this study the DCA is compared against a neural network based Self Organising Map (SOM) approach. Significant statistical differences were found in the performance of the two algorithms, with further one-sided nonparametric statistical tests concluding that the performance of the DCA is superior to that of a standard SOM, when comparing antigen type segment sizes of 10000. For full experimental details and a comprehensive analysis of this comparative study refer to Greensmith *et al.* [Greensmith08a], and [Gu08] for a comparative study of the DCA, negative selection and other machine learning algorithms.

## 5. DCA Development

Numerous stages are involved in creating an immune inspired algorithm. Following the description given in the previous section, here the process is presented by which this algorithm was designed and implemented. This process consisted of numerous stages and commenced by examining the interactions between DCs and T-cells. Once this information was compiled it became apparent that DCs perform a crucial role in mediating between the innate and adaptive immune system.

DCs appear to be a key cell in the immunological decision making process. The model generated at this stage involved multiple signal processing pathways within the DC itself in addition to complex interactions with a variety of adaptive immune cells. This model is highly complicated and is not suitable for direct transformation into an algorithm as it contained too many interactions. An abstract model of this process was required and developed.

The core of the abstract model is shown in Figure 4. This model dictates the cell behaviour and groups multiple cell inputs and outputs into four categories of input signal and three categories of output signal. In this model the state changes of an individual cell are also defined. While a DC is in its signal and antigen collection phase, the cell is termed *immature.* Upon receipt of input signals (PAMPs, danger signals from necrosis and safe signals from apoptosis) the immature cell undergoes a state change to either the *mature* or *semi-mature* state.

Antigen presented by a mature cell are potentially anomalous, and antigen presented by a semi-mature cell are potentially normal. For each type of antigen the number of semi-mature versus mature presentations are counted. This metric is used to derive an anomaly score for that particular type of antigen, upon which a threshold of anomaly is applied. Antigen with a score above this threshold are classed as anomalous. This calculation allows us to dispense with the computationally intensive process of generating T-cells, but provides a similar output functionally.

This abstract model could then be taken and transformed into a feasible algorithm as outlined in Section 3.2. As shown in Figure 6, three incarnations of the algorithm have been developed, implemented and tested on a variety of applications. The initial prototype system provided a ‘proof of concept’, and resulted in a feasible algorithm. This system used the minimum components, using three input signals derived from the dataset, and for each data item, ten artificial DCs were used to sample both antigen (the data ID) and signals (attributes).

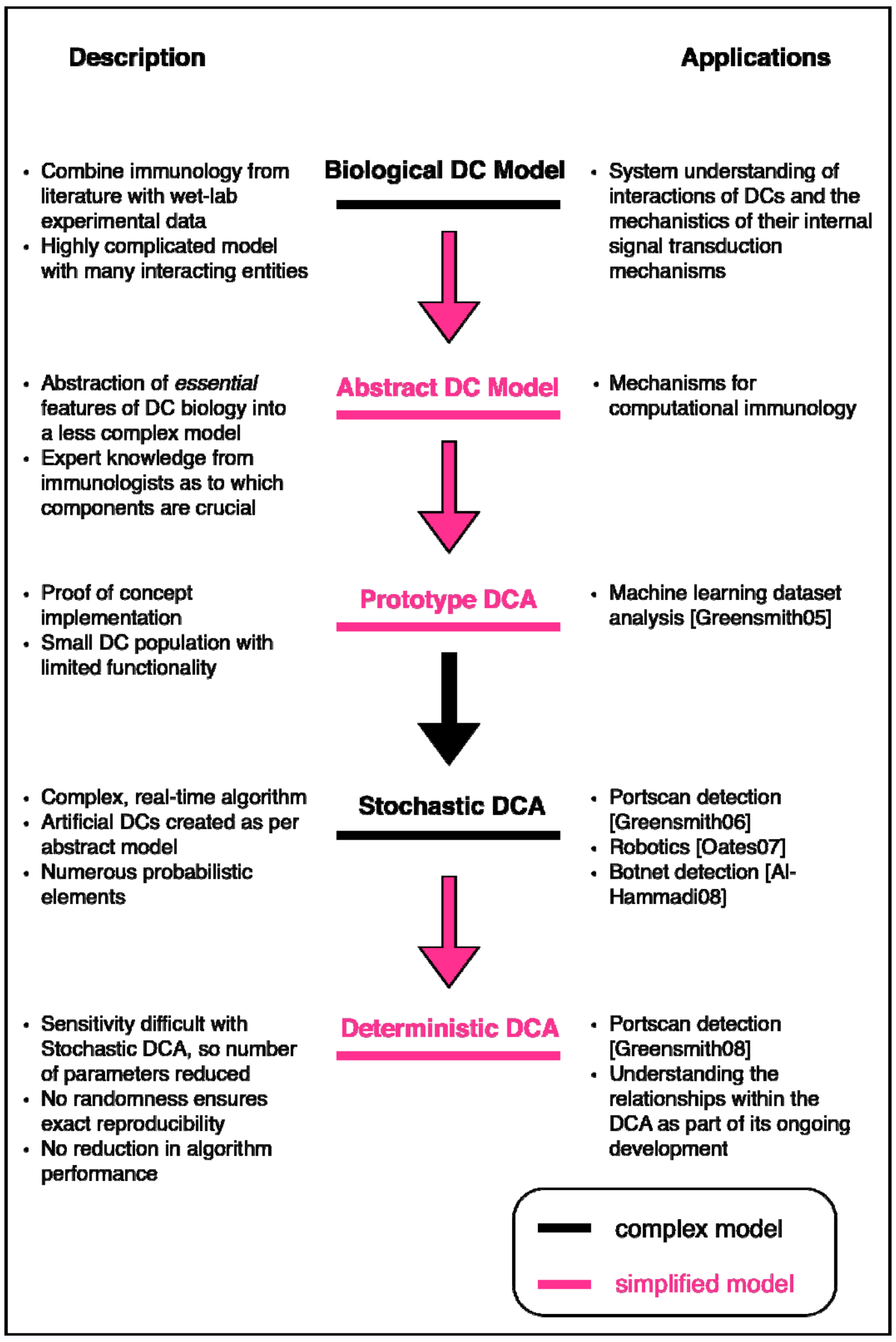

**Figure 6** Diagram of the DCA development process. As shown in the legend, this process cycles between complex and simple models. The appropriate level of complexity is dependent upon the use of the model, shown in the right hand column.

As a rudimentary test, the prototype DCA is applied to the Wisconsin Breast Cancer dataset, where it achieved high rates of true positives and very low rates of false positives. This investigation highlights the suitability of the algorithm for applications which require ordered input data, such as real-time anomaly detection. We demonstrated that the DCA was not suitable for solving standard machine learning problems but could be applied to problems involving intensive processing in a real-time environment.

Following the prototype, the algorithm was scaled up to become a fully working, real-time intrusion detection system [Greensmith06]. An agent-based framework, *libtissue* [Twycross06] was used as a development platform with each DC acting as an independent agent. Antigen are fed into a storage area to be randomly selected at any point by any DC. Signals are fed into a signal matrix, with each member of the DC population updated with new input information each time the matrix is updated. The mechanism used by individual DCs to produce three output signals from this input is explained in the next section. Once all antigen are fed into the system anomaly scores are calculated for each antigen type. To test this system, the DCA is initially applied to the detection of scanning activity from a monitored client machine. As with the proof of concept experiment it is shown that high rates of false positives and low rates of true positives are generated. The initial investigation was then scaled up to encompass more sophisticated scans, where the performance was similarly good. Upon comparison with a Self-Organising Map [Greensmith08a], it is shown that the DCA produces significantly fewer false positives than this established technique.

This particular version of the DCA has also been applied to the detection of a novel threat on the internet, botnets [AlHammadi08], where the DCA produced high rates of true positives and low rates of false positives in comparison to a statistical technique. Outside of computer security Kim *et al.* [Kim06] have successfully applied the DCA to the detection of misbehaviour in wireless sensor networks, where again the algorithm showed much promise. The DCA is also showing promise in the area of robotic security as demonstrated by Oates *et al.* [Oates07]. A proof of concept experiment is performed to demonstrate that the DCA could be used for basic object discrimination in a controlled environment. The same researchers have now extended this research into the theoretical domain [Oates08] through frequency tuning analysis. This research has highlighted that the DCA exhibits filter properties and moreover suggests the importance of the lifespan limit.

We had developed a seemingly successful algorithm capable of good performance across a range of problems and domains. However, this system consisted of over 15 tunable parameters, such as the number of cells, the threshold for maturation, the number of input signals, the weights for the processing of the input signals and numerous other parameters [Greensmith06]. Basic sensitivity analyses could be performed, but was difficult as due to large amounts of probabilistic elements it was not clear which components were performing which function and what exactly was performing the anomaly detection. We suspect that the key to the algorithm is the time-sensitive correlation between processed signals and collected antigen combined with a consensus decision taken across a population of cells. Due to the sheer amount of factors and parameters it was not obvious how we could analyse such a system to the degree of accuracy required.

The most recent incarnation of the DCA is a deterministic version. This remedies the problem of excessive stochastic elements and assists in proving to our community and the bio-inspired community at large not only that the algorithm can produce good results. In addition, we understand why it produces such results. Investigations of the time delay between signals and antigen have assisted in improving our understanding of how the correlation between these two sources of data is performed [Greensmith08].

In addition an improved anomaly assessment and comparable results with our previous systems, it has provided a platform in which we can track individual cells and antigen through the system over numerous repetitions and achieve identical scenarios within our antigen and signal processing. This reproducibility has let us examine the various features in isolation. We aim to extend this work across a multitude of applications and to use it to perform more theoretical analyses of the algorithm. This includes discovering in which situations it is unsuitable along with finding successful applications, allowing for a fuller characterisation of the capabilities of the technique. We intend to use this system as our testbed for adding novel components to the algorithm as the state-of-the-art in immunology progresses including such components as the Th17 cells mentioned previously.

## 6. Conclusions

In this book chapter both human centric and multi-faceted development paradigms have been presented. We have shown the parallels which exist between immunology and artificial immune systems. Such parallels are in terms of development, where immunological discovery has ultimately shaped the way in which we view the immune system in order to construct immune-inspired algorithms. This phenomena may be at least partially attributed to the fact that what is of interest to immunologists is ultimately published and such resources form the basis of inspiration. Perhaps the link between immunology and AIS will become even closer as interdisciplinary collaborations within AIS become more prevalent, resulting in algorithms which actually resemble an immune system. Whether an increased amount of immunological accuracy will be of any great benefit to AIS remains to be seen. However, the close examination of immunology appears to have been fruitful for the DCA.

With the DCA two levels of abstraction were used, namely at an intra-cellular level and at a cell behaviour level. The choice to use these levels in particular was dictated by the scope of experiments performed by the collaborating immunologists. This has resulted in an algorithm which is unique as it performs filtering on input signals, correlation between signals and antigen and classification of antigen. Without such detailed immunology, the inspiration may have appeared too abstract, and the resulting system may have become over simplistic. In order to develop complex algorithms one may need to understand the complex form of the chosen system of inspiration.

Finally, it is crucial to note the manner by which the DCA was developed as shown in Figure 5. This process varied between highly complicated models and systems to simplified versions. The cycling between complex and simple is necessary - the complex models are needed in order to find the correct level of detail, with the simplification process reducing factors such as computational complexity or having to explicitly model interactions between molecules and receptors. Both types of model, simple and complex, are needed in order to find the right level of abstraction to transform an idea into a working system.

The current incarnation, the deterministic DCA, has reduced numbers of parameters and controllable elements, so the same antigen and signals are sampled by the same cell agents provided the input is kept constant. As the simple to complex cycle continues, the next step with this algorithm is to introduce stochastic elements individually. This will allow for the investigation of the algorithm behaviour in more detail, and will assist in demonstrating how much randomness is necessary in this system or similar.

**Acknowledgements:** This work is supported by the EPSRC (EP/D071976/1).